\documentclass[runningheads]{llncs}
\usepackage{graphicx}
\usepackage{comment}
\usepackage{amsmath,amssymb} 
\usepackage{color}

\usepackage{epsfig}
\usepackage{graphicx}
\usepackage{amsmath}
\usepackage{amssymb}
\usepackage{booktabs} 
\usepackage{multirow}
\usepackage{pifont}
\usepackage[font=small,labelfont=bf]{caption}

\newcommand{\cmark}{\ding{51}}%
\newcommand{\xmark}{\ding{55}}%

\begin{document}
\pagestyle{headings}
\mainmatter
\def\ECCVSubNumber{2506}  

\title{MetAdapt: Meta-Learned Task-Adaptive Architecture for Few-Shot Classification} 

\titlerunning{MetAdapt: Meta-Learned Task-Adaptive Architecture for Few-Shot}
%
\author{Sivan Doveh*$^{1,2}$,
Eli Schwartz*$^{1,2}$, 
Chao Xue$^{1}$, 
Rogerio Feris$^{1}$,
\\
Alex Bronstein$^{3}$, 
Raja Giryes$^{2}$, 
Leonid Karlinsky$^{1}$
}
\authorrunning{S. Doveh et al.}
%
\institute{$^{1}$ IBM Research AI, 
$^{2}$ Tel Aviv University, Israel, 
$^{3}$ Technion, Israel}
\maketitle

\let\thefootnote\relax\footnotetext{\small{* Equal contributors
\\Corresponding authors:
Sivan Doveh \texttt{sivan.doveh@ibm.com} and Leonid Karlinsky \texttt{leonidka@il.ibm.com}}}

\begin{abstract}
Few-Shot Learning (FSL) is a topic of rapidly growing interest. Typically, in FSL a model is trained on a dataset consisting of many small tasks (meta-tasks) and learns to adapt to novel tasks that it will encounter during test time. This is also referred to as meta-learning. Another topic closely related to meta-learning with a lot of interest in the community is Neural Architecture Search (NAS), automatically finding optimal architecture instead of engineering it manually. In this work we combine these two aspects of meta-learning.
So far, meta-learning FSL methods have focused on optimizing parameters of pre-defined network architectures, in order to make them easily adaptable to novel tasks. Moreover, it was observed that, in general, larger architectures perform better than smaller ones up to a certain saturation point (where they start to degrade due to over-fitting). However, little attention has been given to explicitly optimizing the architectures for FSL, nor to an adaptation of the architecture at test time to particular novel tasks. In this work, we propose to employ tools inspired by the Differentiable Neural Architecture Search (D-NAS) literature in order to optimize the architecture for FSL without over-fitting. Additionally, to make the architecture task adaptive, we propose the concept of `MetAdapt Controller' modules. These modules are added to the model and are meta-trained to predict the optimal network connections for a given novel task. Using the proposed approach we observe state-of-the-art results on two popular few-shot benchmarks: \textit{mini}ImageNet and FC100.
\vspace{-0.1cm}
%

\end{abstract}

\section{Introduction}

\begin{figure*}[!htb]
  \centering
\includegraphics[width=1.1\linewidth]{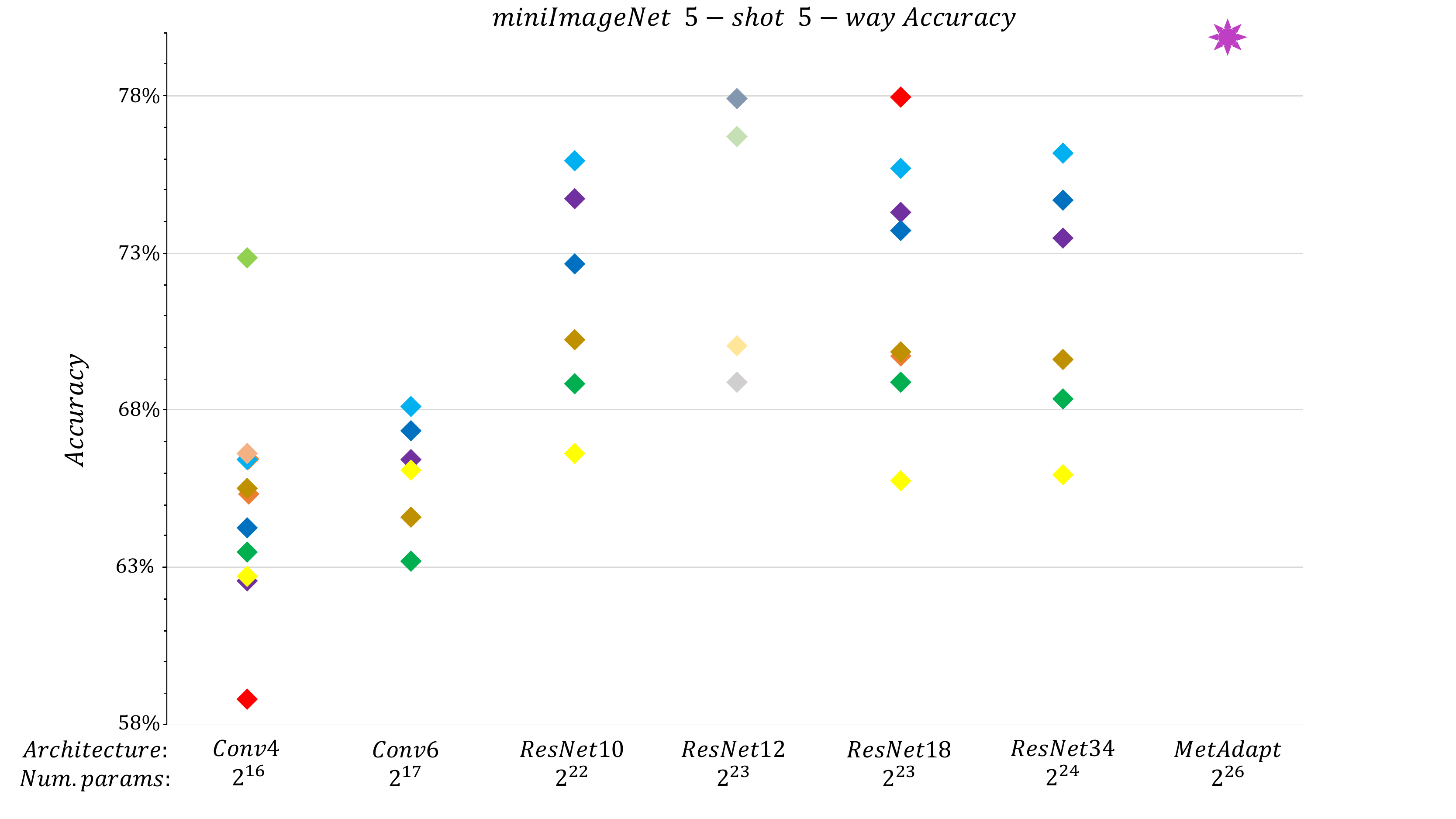}
  \caption{\textbf{Accuracy vs. Architecture} as can be seen, the general trend is that larger architectures improve performance. However, when reaching a certain size (ResNet18/34) performance saturates or even degrades due to over-fitting. We propose MetAdapt as a method for finding and training larger models that still improve performance. Markers represent results of different methods with various backbone architectures for the 5-shot / 5-way miniImageNet few-shot benchmark. Same-color markers indicate same method with different architectures. Exact numbers and references for the data points in the figure are provided in the supplementary material.}
  \label{fig:acc_vs_arch}
\end{figure*}

Recently, there has been a lot of exciting progress in the field of few-shot learning in general, and in few-shot classification (FSC) in particular. 
A popular method for approaching FSC is meta-learning, or learning-to-learn. In meta-learning, the inputs to both train and test phases are not images, but instead a set of few-shot tasks, $\{T_i\}$, each $K$-shot / $N$-way task containing a small amount $K$ (usually 1-5) of labeled support images and some amount of unlabeled query images for each of the $N$ categories of the task. The goal of meta-learning is to find a base model that is easily adapted to the specific task at hand, so that it will generalize well to tasks built from novel unseen categories and fulfill the goal of FSC (see Section \ref{related} for further review).

Many successful meta-learning based approaches have been developed for FSC \cite{Vinyals2016,Snell2017,Finn2017,nichol2018first,leo,tadam,lee2019meta} advancing its state-of-the-art. Besides continuous improvements offered by the FSC methods, some general trends affecting the performance of FSC have become apparent. One of such major factors is the CNN backbone architecture at the basis of all the modern FSC methods.
Carefully reviewing and placing on a single chart the test accuracies of top-performing FSC approaches w.r.t. the backbone architecture employed reveals an interesting trend (Figure \ref{fig:acc_vs_arch}).
It is apparent that larger architectures increase FSC performance, up to a certain size, where performance seems to saturate or even degrade.
This happens since bigger backbones carry higher risk of over-fitting. It seems the overall performance of the FSC techniques cannot continue to grow by simply expanding the backbone size.

In light of the above, in this paper we set to explore methods for architecture search, their meta-adaptation and optimization for FSC. Neural Architecture Search (NAS) is a very active research field that has contributed significantly to overall improvement of the state of the art in supervised classification. Some of the recent NAS techniques, and in particular Differentiable-NAS (D-NAS), such as DARTS \cite{liu19darts}, are capable of finding optimal (and transferable) architectures given a particular task using a single GPU in the course of 1-2 days. This is due to incorporating the architecture as an additional set of neural network parameters to be optimized, and solving this optimization using SGD. Due to this use of additional architectural parameters, the training tends to over-fit. D-NAS optimization techniques are especially designed to mitigate over-fitting, making them attractive to extreme situations with the greatest risk of over-fitting, such as in the case of FSC. 

So far, D-NAS techniques have been explored mainly in the context of large scale tasks, involving thousands of labeled examples for each class. Very little work has been done on NAS for few-shot. D-NAS in particular, to the best of our knowledge, has not been applied to few-shot problems yet. Meta-adaption of the architecture in task dependent manner to accommodate for novel tasks also has not been explored.

In this work, we build our few-shot task-adaptive architecture search upon a technique from D-NAS (DARTS \cite{liu19darts}). Our goal is to learn a neural network where connections are controllable and adapt to the few-shot task with novel categories. Similarly to DARTS, we have a neural network in the form of a Directed Acyclic Graph (DAG), where the nodes are the intermediate feature maps tensors, and edges are operations. Each edge is a weighted sum of operations (with weights summing to 1), each operation is a different preset sequence of layers (convolution, pooling, BatchNorm and non-linearity). The operations set includes the identity-operation and the zero-operation to either keep the representation untouched or cut the connection. To avoid over-fitting, a bi-level (two-fold) optimization is performed where first the operation layers' weights are trained on one fold of the data and then the connections' weights are trained on the other fold. 

However, unlike DARTS, our goal is not to learn a one time architecture to be used for all tasks. To be successful at FSC, we need to make our architecture task adaptive so it would be able to quickly rewire for each new target task. To this end, we employ a set of small neural networks, MetAdapt Controllers, responsible for controlling the connections in the DAG given the current task. The MetAdapt Controllers adjust the weights of the different operations, such that if some operations are better for the current task they will get higher weights, thus, effectively modifying the architecture and adapting it to the task.

To summarize, our contributions in this work are as follows: (1) We show that DARTS-like bi-level iterative optimization of layer weights and network connections performs well for few-shot classification without suffering from overfitting due to over-parameterization; (2) We show that adding small neural networks, MetAdapt Controllers, that adapt the connections in the main network according to the given task further (and significantly) improves performance; (3) using the proposed method, we obtain improvements over FSC state-of-the-art on two popular FSC benchmarks: \textit{mini}ImageNet \cite{Vinyals2016} and FC100 \cite{tadam}.


\section{Related Work}
\label{related}

{\bf Few-Shot Learning.} 
The major approaches to few-shot learning include: metric learning, generative (or augmentation) based methods, and meta learning (or learning-to-learn).

\paragraph{Few-shot learning by metric learning.}
This type of methods \cite{Weinberger2009,Snell2017,Rippel2015} learn a non-linear embedding into a metric space where $L_2$ nearest neighbor (or similar) approach is used to classify instances of new categories according to their proximity to the few labeled training examples embedded in the same space. Additional proposed variants include using a metric learning method based on graph neural networks \cite{Garcia2017}, that goes beyond the $L_2$ metric. Similarly, \cite{Santoro2016,relationnet} introduce metric learning methods where the similarity is computed by an implicit learned function rather than via the $L_2$ metric over an embedding space.

The embedding space based metric-learning approaches are either posed as a general discriminative distance metric learning \cite{Rippel2015,chopra05}, or optimized on the few-shot tasks \cite{Snell2017,Weinberger2009,Garcia2017,tadam}, via the meta-learning paradigm that will be described next. These approaches show a great promise, and in some cases are able to learn embedding spaces with some meaningful semantics embedded in the metric \cite{Rippel2015}. Improved performance in the metric learning based methods has been achieved when combined with some additional semantic information. In \cite{caml}, class conditioned embedding is used. In \cite{am3}, the visual prototypes are refined using a corresponding label embedding and in \cite{schwartz2019baby} it is extended to using multiple semantics, such as textual descriptions.

\paragraph{Augmentation-based few-shot learning.}
This family of approaches refers to methods that (learn to) generate more samples from the one or a few examples available for training in a given few-shot learning task. These methods include synthesizing new data from few examples using a generative model, or using external data for obtaining additional examples that facilitate learning on a given few shot task. These approaches include: (i) semi-supervised approaches using additional unlabeled data \cite{Dong2017,Fu2015}; (ii) fine tuning from pre-trained models \cite{Li2016,Wang2016a,Wang2016}; (iii) applying domain transfer by borrowing examples from relevant categories \cite{Lim2012} or using semantic vocabularies \cite{Ba2015,Fu2016a}; (iv) rendering synthetic examples  \cite{Park2015,Dosovitskiy2017,Su2015}; (v) augmenting the training examples using geometric and photometric transformations \cite{Krizhevsky2012} or learning adaptive augmentation strategies \cite{Guu2017}; (vi) example synthesis using Generative Adversarial Networks (GANs) \cite{Zhu2017,Zhu2016,Goodfellow2014,Reed2018,Radford2015,Mao2016,Durugkar2017,Huang2017,Antoniou2018}.

In \cite{Hariharan2017,Schwartz2018} additional examples are synthesized via extracting, encoding, and transferring to the novel category instances, of the intra-class relations between pairs of instances of reference categories. In \cite{Wang2018}, a generator sub-net is added to a classifier network and is trained to synthesize new examples on the fly in order to improve the classifier performance when being fine-tuned on a novel (few-shot) task. In \cite{Reed2018}, a few-shot class density estimation is performed with an auto-regressive model, augmented with an attention mechanism, where examples are synthesized by a sequential process. In \cite{Chen2018,Yu2017} label and attribute semantics are used as additional information for training an example synthesis network. In \cite{alfassy2019laso} models are trained to perform set-operations (e.g. union) and then can be used to synthesise samples for few-shot multi-label classifications. 

\paragraph{Few-shot meta-learning (learning-to-learn).}
These methods are trained on a set of few-shot tasks (also known as `episodes') instead of a set of object instances, with the motivation to learn a learning strategy that will allow effective adaptation to new such (few-shot) tasks using one or few examples.

An important sub-category of meta learning methods is metric-meta-learning, combining metric learning as explained above with task-based (episodic) training of meta-learning. In Matching Networks \cite{Vinyals2016}, a non-parametric $k$-NN classifier is meta-learned such that for each few-shot task the learned model generates an adaptive embedding space for which the task can be better solved. In \cite{Snell2017} the metric (embedding) space is optimized such that in the resulting space different categories form compact and well separated uni-modal distributions around the category `prototypes' (centers of the category modes).

Another family of meta-learning approaches is the so-called `gradient based approaches', that try to maximize the `adaptability', or speed of convergence, of the networks they train to new (few-shot) tasks (usually assuming an SGD optimizer). In other words, the meta-learned classifiers are optimized to be easily fine-tuned on new few-shot tasks using small training data. The first of these approaches is MAML \cite{Finn2017} that due to its universality was later extended through many works such as, Meta-SGD \cite{Li2017}, DEML+Meta-SGD \cite{Zhou2018}, Meta-Learn LSTM \cite{Ravi2017}, and Meta-Networks \cite{Munkhdalai2017}. In LEO \cite{leo}, a MAML like loss is applied not directly on the model parameters, but rather on a latent representation encoding them. This approach featured an encoder and a decoder to and from that latent space and achieved state-of-the-art results on \textit{mini}Imagenet few-shot benchmark among models relying on visual information alone.

In MetaOptNet \cite{lee2019meta} a CNN backbone is trained end-to-end with an unrolled convex optimization solution of an optimal classifier, such as SVM. In this work, we use their suggested construction of performing SGD through an unrolled SVM optimization to train the backbone. Our work is focused on optimizing the backbone architecture.

Other methods focus on regularization for mitigating the over-fitting. In BF3S \cite{gidaris2019boosting} auxiliary self-supervision tasks are added, such as predicting image rotation or patch location. In Robust-dist \cite{Dvornik_2019_ICCV} first an ensemble of up to 20 models is learned, so each model by itself cannot overfit the data. Then, the the final model is distilled from all those models. Notably, our method which also deals with training large architecture without over-fitting is orthogonal to these two approaches. It is likely that further improvement can be achieved by combining these methods with ours. 

Notably, in all previous meta-learning methods, only the parameters of a (fixed) neural network are optimized through meta-learning in order to become adaptable (or partially adaptable) to novel few-shot tasks. In this work, we both learn a specialized backbone architecture that would facilitate this adaptability, as well as meta-learn to become capable of adapting that architecture itself to the task, thus going beyond the parameter only adaptation of all previous meta-learning approaches.

{\bf Neural Architecture Search.}
Over the last few years Neural Architecture Search (NAS) have enabled automatic design of novel architectures that outperformed previous hand-designed architectures in terms of accuracy.
Two notable works on NAS are AmoebaNet \cite{Real18Regularized} and NASnet \cite{zophNasRL}.
The first one used a genetic algorithm and the second used a reinforcement learning based method. Although achieving state of the art performance at the time, these methods required enormous amount of GPU-hours . Efficient NAS (ENAS) \cite{ENAS}, a reinforcement learning based method, used weight sharing across its child models, which are sub graphs of a larger one.
By that, they managed to accelerate the search process. The work in \cite{effnet} shows how to scale the size of such learned architectures with the size of the input data.

Recently, differentiable methods with lower demand for computing have been introduced. Notable among them are differentiable architecture search (DARTS) \cite{liu19darts} and SNAS \cite{snas}. These methods managed to search for architecture in just a few GPU days.
DARTS relaxes the search space into a continuous one, allowing a differentiable search.
The DARTS method includes two stages. First, basic structures are learned, by placing coefficients attached to operations between feature maps. These coefficients indicate the importance of the attached operations and connections. After the search is done, the final basic structures are formed by pruning and keeping only the most important operations.
At the second stage, the final network is built by repeatedly concatenating the found basic structures.

ASAP \cite{ASAP} addresses the issue that harsh pruning at the end of the search makes the found architecture sub-optimal. It does so by performing gradual pruning. ASAP achieves higher accuracy with a shorter training time.
In SNAS \cite{snas}, the search is done by learning a continuous architectures distribution and sampling from it. This distribution is pushed closer to binary by using a temperature parameter and gradually decreasing it.
Then, the chosen architecture is the one that has the higher probability. In \cite{proxy} a binary mask is learned and used to keep a single path of the network graph. By doing so, they managed to search for the whole network and not only cells.
PNAS \cite{liu2018progressive} suggested a method for progressively searching for a larger architecture. P-DARTS \cite{chen2019progressive} do the same but with differentiable architecture search.


These methods are mostly focused, and perform well, on datasets such as CIFAR and ImageNet. So far, little attention has been given to their adaptation to few-shot learning. Auto-Meta \cite{kim2018auto} used PNAS \cite{liu2018progressive} based search for few-shot classification, but with a focus on searching for a small architecture (resulting in a relatively low performance w.r.t. to current state-of-the-art).  In particular, the possibility of adapting the architecture at test-time to a specific novel task, as proposed in this work, has not been explored before.

\section{MetAdapt}

\begin{figure*}[tbh]
  \centering
\includegraphics[width=1\linewidth]{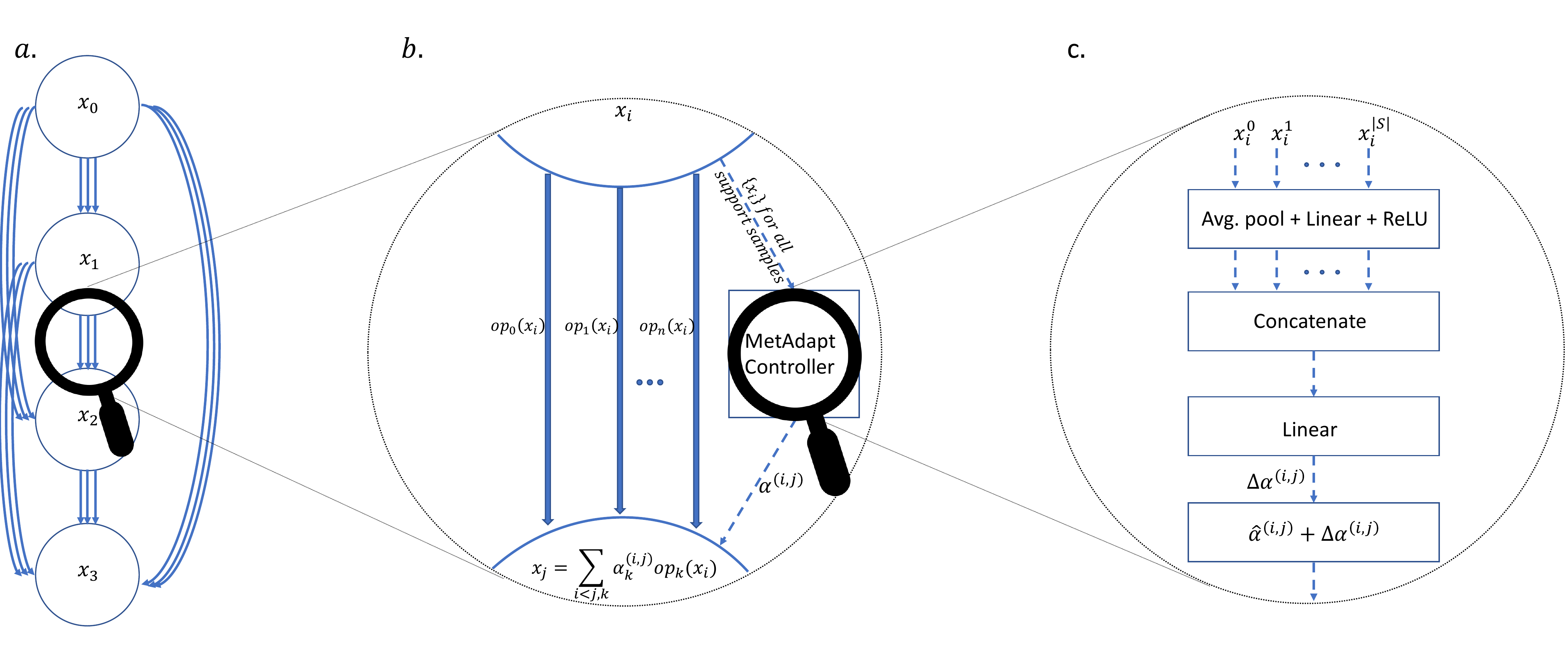}
  \caption{\textbf{MetAdapt Block Architecture Illustration.} MetAdapt block has a DAG structure for the network with edges being operations (network layers) and nodes being the feature maps calculated as a convex combination of all edges arriving at a node. The weighting of each edge is optimized in a bi-level manner in parallel with the layer parameters. Additionally, MetAdapt controllers receive as input the concatenated average pooled feature grids for  all the support-samples of the current episode and adapts the edges' convex combination weights according to the current task.}
  \label{fig:block_arch}
\end{figure*}

In this section we describe the architecture and training procedure for MetAdapt.
We introduce the task-adaptable block, it has a graph structure with adaptable connections that can modulate the architecture, adapting it to the few-shot task at hand. We then describe the sub-models, MetAdapt Controllers, that predict the change in connectivity that is needed in the learned graph as a function of the current task. Finally, we describe the training procedure.


\subsection{Task-Adaptable Block}
The architecture of the adaptable block used in MetAdapt is defined, similarly to DARTS \cite{liu19darts}, as a Directed Acyclic Graph (DAG).
The block is built from feature maps $V=\{x_i\}$ that are linked by mixtures of operations. The input feature map to the block is $x_0$ and its output is $x_{|V|-1}$.
A \emph{Mixed Operation}, $\bar{o}^{(i,j)}$, is defined as
\begin{equation}\label{eq:mixedOp}
\bar{o}^{(i,j)}(x)=\frac{\sum_{o\in\mathcal{O}}exp(\alpha_o^{(i,j)})o(x)}{\sum_{o\in\mathcal{O}}exp(\alpha_{o}^{(i,j)})},
\end{equation}
where $\mathcal{O}$ is a set of the search space operations, $o(x)$ is an operation applied to $x$, and $\alpha_o^{(i,j)}$ is an optimised coefficient for operation $o$ at edge $(i,j)$. Later, we will describe how $\alpha$s can be adapted per task ($K$-shot, $N$-way episode). The list of search space operations used in our experiments is provided in Table \ref{tab:operations}. This list includes the zero-operation and identity-operation that can fully or partially (depending on the corresponding $\alpha_{o}^{(i,j)}$) cut the connection or make it a residual one (skip-connection). Each feature map $x_i$ in the block is connected to all previous maps by setting it to be:
\begin{equation}
x_i = \sum_{j<i}\bar{o}^{(j,i)}(x_j).
\end{equation}
The task-adaptive block defined above can be appended to any backbone feature extractor. Potentially, more than one block can be used.
We use ResNet9 followed by a single task-adaptive block with $4$ nodes ($|V|=4$) in our experiments, resulting in about $8$ times more parameters compared with the original ResNet12 (due to large set of operations on all connections combined). Note that as we use $4$ nodes in our block, there exists a single path in our search space that is the regular residual block (ResNet3 block).
Figure \ref{fig:block_arch}a schematically illustrates the block architecture.

\definecolor{c0}{rgb}{0.9,0.17,0.31} 
\definecolor{c1}{rgb}{0.95,0.61,0.73}
\definecolor{c2}{rgb}{0.71,0.2,0.54}
\definecolor{c3}{rgb}{0.6,0.4,0.8} 
\definecolor{c4}{rgb}{0.3,0.1,0.5}
\definecolor{c5}{rgb}{0.0,0.53,0.74}
\definecolor{c6}{rgb}{0.0,0.72,0.92} 
\definecolor{c7}{rgb}{0.67,0.88,0.69} 
\definecolor{c8}{rgb}{0.18,0.52,0.49} 

\begin{minipage}{0.96\textwidth}
\hspace{-0.6cm}
    \begin{minipage}[c]{0.5\textwidth}
        \includegraphics[width=1\linewidth]{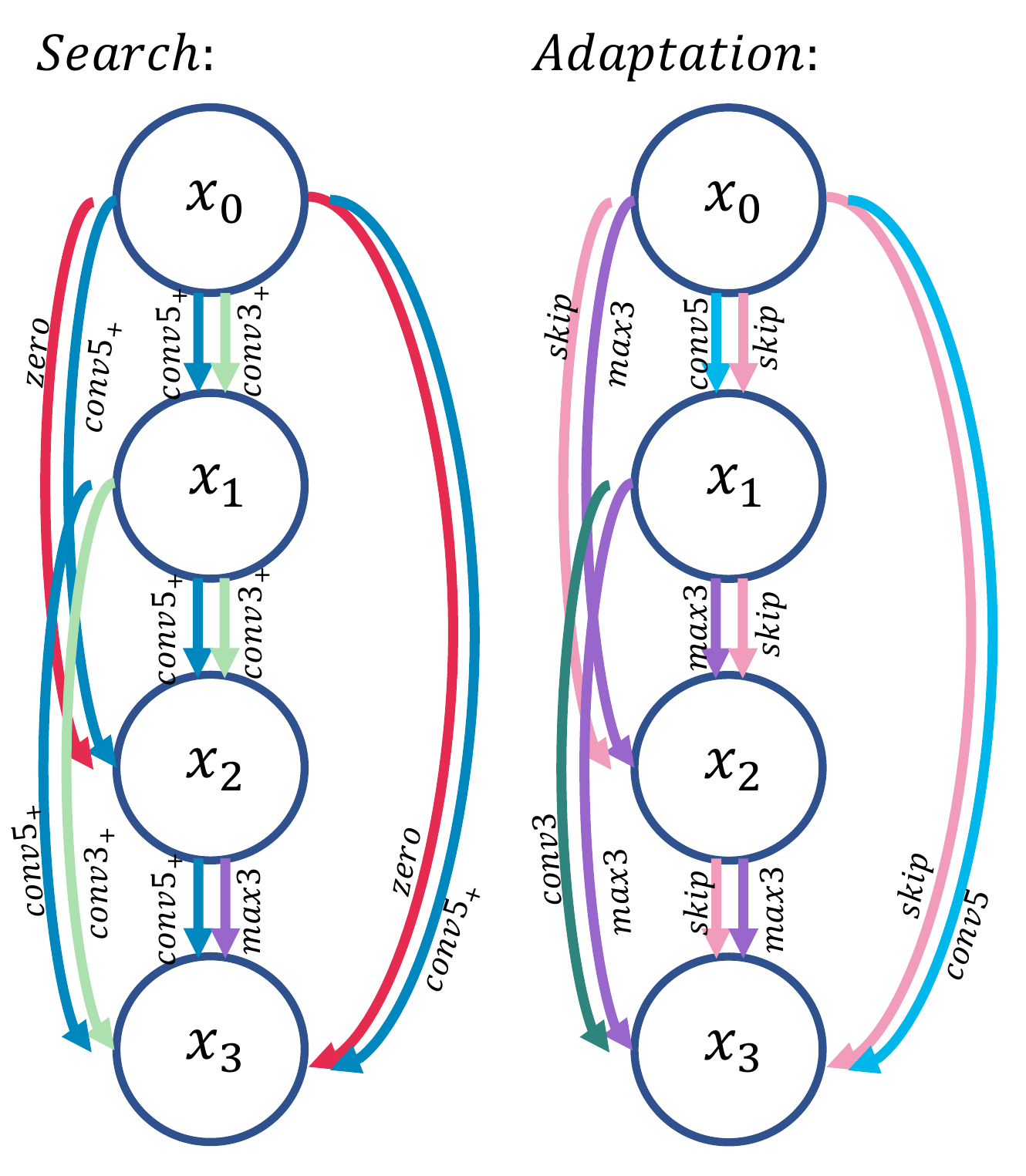}
  \captionof{figure}{\textbf{Optimized architecture visualization} For each edge the top-2 operations are visualized, color-coded according to Table \ref{tab:operations}. On the left are the top operations after search ($\hat{\alpha}$). On the right are the operations predicted by `MetAdapt Controllers' to be the most important for a specific random task and got the highest extra weighting ($\Delta\alpha$).}
  \label{fig:episodic_adapt_dag}
    \end{minipage}
    \hfill
    \begin{minipage}[c]{0.5\textwidth}
    \begin{minipage}{1\textwidth}
        \captionof{table}{\textbf{List of possible \\ operations on each edge} }
        \begin{small}
        \begin{tabular}{ll}
            \toprule
            Op ID & layers \\
            \midrule
            \color{c0}{$zero$} & \color{c0}{The zero op. - cut connection} \\
            \color{c1}{$skip$} & \color{c1}{The identity op. - skip connect} \\
            \color{c2}{$mean3$} & \color{c2}{Average Pool $3\times3$ $\rightarrow$ BN} \\
            \color{c3}{$max3$} & \color{c3}{Max Pool $3\times3$ $\rightarrow$ BN} \\
            \color{c4}{$conv1$} & \color{c4}{Conv $1\times1$ $\rightarrow$ BN} \\
            \color{c5}{$conv5_+$} & \color{c5}{Conv $5\times5$ $\rightarrow$ BN $\rightarrow$} \\ 
            & \color{c5}{LeakyReLU(0.1)} \\
            \color{c6}{$conv5$} & \color{c6}{Conv $5\times5$ $\rightarrow$ BN} \\
            \color{c7}{$conv3_+$} & \color{c7}{Conv $3\times3$ $\rightarrow$ BN $\rightarrow$} \\
            & \color{c7}{LeakyReLU(0.1)} \\
            \color{c8}{$conv3$} & \color{c8}{Conv $3\times3$ $\rightarrow$ BN} \\
            
            \bottomrule
        \end{tabular}
        \end{small}
        \label{tab:operations}
        \end{minipage}
        
        \begin{minipage}{1\textwidth}
        \includegraphics[width=1\textwidth]{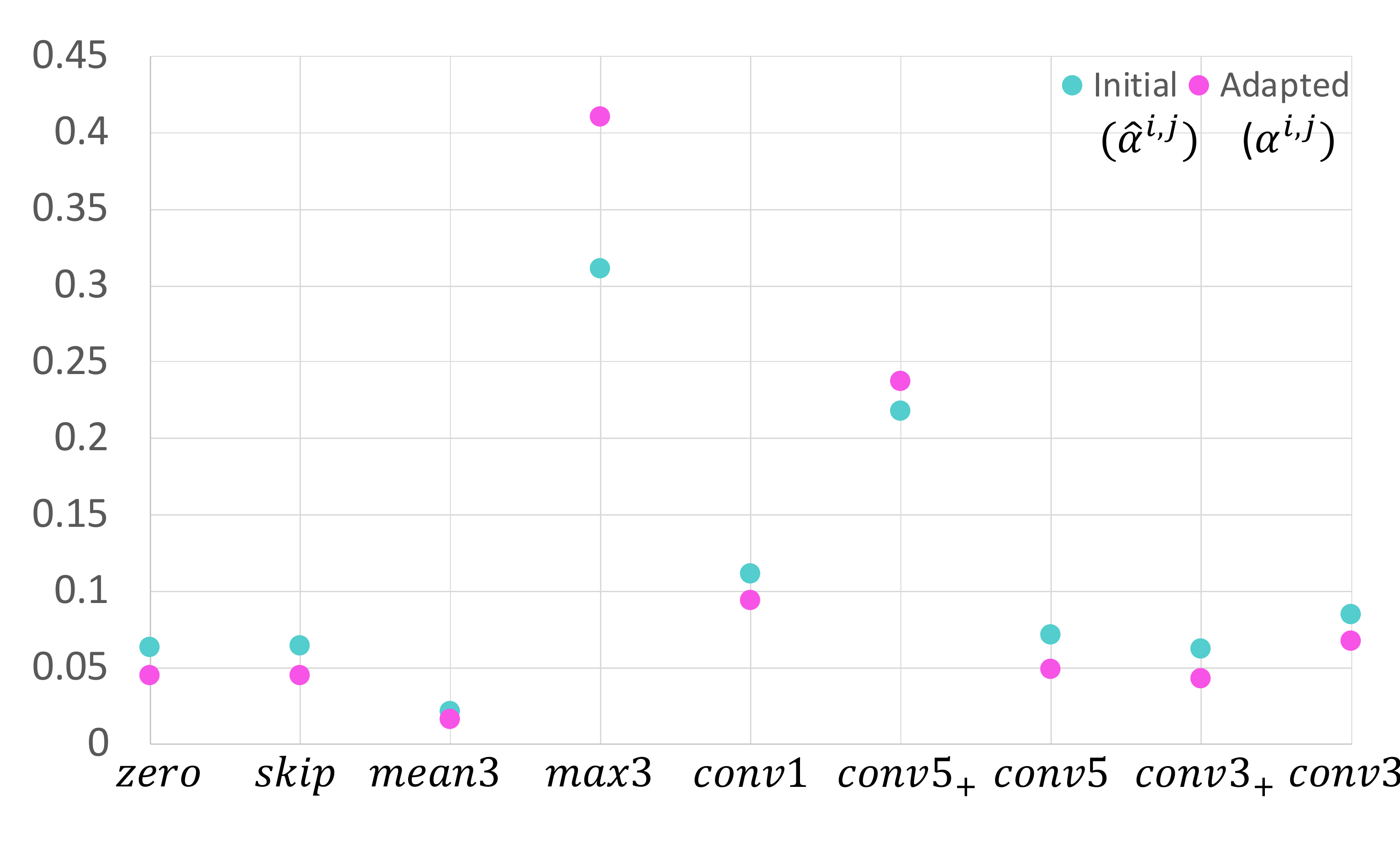}
          \captionof{figure}{Example of global $\hat{\alpha}^{(i,j)}$ and $\alpha^{(i,j)}$ after the adaptation made by MetAdapt Controller for a specific few-shot episode}
          \label{fig:episodic_adapt}
        \end{minipage}
    \end{minipage}
\end{minipage}

\subsection{MetAdapt Controllers}
The task-adaptive block is accompanied by a set of MetAdapt Controller modules, one per edge. They are responsible for predicting, given a few-shot task, the best way of adapting the mixing coefficients ($\alpha_o^{(i,j)}$) for the corresponding edge operations. Let $\alpha^{(i,j)}$ be the vector of all $\alpha_o^{(i,j)}$. Let $\hat{\alpha}^{(i,j)}$ be the globally optimized coefficients (optimization process described below), then MetAdapt controllers predict the task-specific residuals $\Delta \alpha^{(i,j)}$, that is the modification required to make to $\hat{\alpha}^{(i,j)}$ for the current task (few-shot episode). Finally,
\begin{equation}
\alpha^{(i,j)} = \hat{\alpha}^{(i,j)}+\Delta \alpha^{(i,j)}
\end{equation}
are the final task-adapted coefficients used for the \emph{Mixed Operation} calculation as in Equation \ref{eq:mixedOp}.

The architecture for each MetAdapt Controller, predicting $\Delta \alpha^{(i,j)}$, is as follows. It receives the input feature maps of the corresponding edge $x_i$, computed for all the support samples of the episode. For a support-set of size $S$, number of channels $D$ and feature map spatial resolution $M\times M$, the input is a tensor of dimensions $(S,D,M,M)$. We perform global average pooling to obtain a $(S,D)$ tensor, followed by
a bottleneck linear layer (with ReLU activation) that operates on each sample individually, to get a $(S,D_{bottleneck})$ size tensor. Then, all support samples representations are concatenated to form a single vector of size $S \cdot D_{bottleneck}$. Finally, another linear layer
maps the concatenation of all support-samples to the predicted $\Delta \alpha^{(i,j)}$. The MetAdapt controller architecture and the way it is used in our adaptable block structure are schematically illustrated on Figure \ref{fig:block_arch}b+c. Figures \ref{fig:episodic_adapt} and \ref{fig:episodic_adapt_dag} present an example of adaptation made by the MetAdapt Controller for a specific episode.

\subsection{Training}
Replacing simple sequence of convolutional layers with the suggested DAG, with its many layers and parameters, in conventional gradient descent training will result in a larger over-fitting. This is even worse for FSL, where it is harder to achieve generalization due to scarcity of the data and the domain differences between the training and test sets. Researchers have faced the same problem with differentiable architecture search, where the objective is to train a large neural network with weighted connections that are then pruned to form the final chosen architecture. 

We follow the solution proposed in DARTS \cite{liu19darts}, solving a bi-level iterative optimization of the layers' weights $w$ and the coefficients of operations $\alpha$ between the nodes. The training set is split to $train_{w}$ for weights training and $train_{\alpha}$ for training the $\alpha$'s. Iteratively optimizing $w$ and $\alpha$ to convergence is prohibitively slow. So, like in DARTS, $w$ is optimized with a standard SGD:
\begin{equation}\label{eq:w_update}
    w = w - \mu \nabla_w Loss_{train_{w}}(w,\alpha),
\end{equation}
where $\mu$ is the learning rate. The $\alpha$'s are optimized using SGD with a second-order approximation of the model after convergence of $w$, by applying:
\begin{equation}\label{eq:alpha_update}
    \alpha = \alpha  - \eta \nabla_\alpha Loss_{train_{\alpha}}(w - \mu Loss_{train_{w}}(w,\alpha),\alpha)
\end{equation}
where $\eta$ is the learning rate for $\alpha$.
The MetAdapt Controllers' parameters are trained as a final step, with all other parameters freezed, using SGD on the entire training set for a single epoch.

\begin{table}[tbh]
\caption{\textbf{miniImageNet 5-way accuracy.} $transferred$ means the architecture was searched on FC100 and transferred to minImageNet}
\begin{center}
\begin{tabular}{llcc}
\toprule
& & \multicolumn{2}{c}{miniImageNet} \\
Method & Architecture & 1-shot & 5-shot \\
\midrule
Matching Networks \cite{Vinyals2016,closer_look} & ResNet10 & 54.49 & 68.82 \\
MAML \cite{Finn2017,closer_look} & ResNet10 & 54.69 & 66.62  \\
ProtoNet \cite{Snell2017,closer_look} & ResNet18 & 54.16 & 73.68 \\
RelationNet \cite{Sung2017LearningLearning,closer_look} & ResNet18 & 52.48 & 69.83 \\
Auto-Meta \cite{kim2018auto} & -- & 51.16 & 69.18 \\
Baseline \cite{closer_look} & ResNet10 & 52.37 & 74.69 \\
Baseline++ \cite{closer_look} & ResNet10 & 53.97 & 76.16 \\
SNAIL \cite{Mishra2017AMeta-Learner} & ResNet12 & 55.71 & 68.88 \\
Dynamic Few-shot \cite{gidaris2018dynamic} & WResNet28 & 56.20 & 73.00 \\
AdaResNet \cite{munkhdalai2018rapid} & -- & 56.88 & 71.94 \\
TADAM \cite{tadam} & ResNet12 & 58.50 & 76.70   \\
Activation to Parameter \cite{qiao2018few} & WResNet28 & 59.60 & 73.74 \\
$\Delta$-Encoder \cite{Schwartz2018} & ResNet18 & 59.90 & 69.70 \\
wDAE \cite{gidaris2019generating} & WResNet28 & 61.07 & 76.75  \\
LEO \cite{leo} & WResNet28 & 61.76 & 77.52 \\
MetaOptNet \cite{lee2019meta} & ResNet12 & 62.64 & 78.63  \\
BF3S \cite{gidaris2019boosting} & WResNet28 & 62.93 & 79.87 \\
Robust-dist \cite{Dvornik_2019_ICCV} & ResNet18 & 63.06 & \textbf{80.63}\\
\midrule
$MetAdapt_{transferred}$ (Ours) &  DAG & 62.82 & 79.35 \\
$MetAdapt$ (Ours) & DAG & \textbf{64.80} & \textbf{80.64} \\
\bottomrule
\end{tabular}
\end{center}
\label{tab:results-mini}
\end{table}

\section{Experiments}

\paragraph{Datasets.}
We use the popular \textit{mini}ImageNet and FC100 few-shot benchmarks to evaluate our method.

The \textbf{\textit{mini}ImageNet dataset} \cite{Vinyals2016} is a standard benchmark for few-shot image classification, that has $100$ randomly chosen classes from ILSVRC-2012 \cite{imagenet}. These classes are randomly split into $64$ meta-training, $16$ meta-validation, and $20$ meta-testing classes. Each class has 600 images of size $84 \times 84$. We use the same classes splits as \cite{lee2019meta} and prior works.

The \textbf{FC100 dataset} \cite{tadam} is constructed from the  CIFAR-100 dataset \cite{Krizhevsky2009}, which contains $100$ classes that are grouped into $20$ super-classes. These are in turn partitioned into $60$ classes from $12$ super-classes for meta-training, $20$ classes from $4$ super-classes for meta-validation, and $20$ classes from $4$ super-classes for meta-testing. This minimizes the semantic overlap between classes of different splits. Each class contains $600$ images of size $32 \times 32$.

\subsection{Implementation Details}
\label{sub:impl_details}
We use the SVM classifier head as suggested in MetaOptNet \cite{lee2019meta}.
We begin with training a ResNet12 backbone on the training set of the relevant dataset for 60 epochs.
We then replace the last residual block of the ResNet12 backbone with our DAG task-adaptive block, keeping the first 3 ResNet blocks (ResNet9) fixed and perform the architecture search for 10 epochs. Finally, we train the `MetAdapt controllers' for a single epoch. Each epoch consists of 8000 episodes with mini-batch of 4 episodes.

For the initial training we use the SGD optimizer with $intial\ learning\ rate=0.1$, $momentum = 0.9$ and $weight\ decay = 5 \cdot 10^{-4}$. Decreasing the learning rate to 0.006 at epoch 20, 0.0012 at epoch 40 and 0.00024 at epoch 50. 
For weights optimization during the search and meta adaptation phases we use the SGD optimizer with $learning\ rate=0.001$, $momentum = 0.9$ and $weight\ decay = 5 \cdot 10^{-4}$.
For the architecture optimization we use Adam optimizer with $learning\ rate=3\cdot 10^{-4}$, $\beta = \left[0.5,0.99\right]$, $weight\ decay = 10^{-3}$ and the Cosine Annealing learning rate scheduler with $\eta_{min} = 0.004$.

Following previous works, e.g. \cite{Finn2017,closer_look}, we perform test time augmentations and fine-tuning. We perform horizontal flip augmentation, effectively doubling the number of support-set. We fine-tune the DAG weights for 10 iterations where the horizontally flipped support set serves as our labeled query set.

\subsection{Results}

Tables \ref{tab:results-mini} and \ref{tab:results-fc100} compare the MetAdapt performance with the state-of-the-art few-shot classification methods that use plain ResNet backbones. We observe improved results for FC100 1-shot ($+3.46\%$) and 5-shot ($+3.17\%$) and also for miniImageNet 1-shot ($+1.74\%$) and similar results for 5-shot. We see that despite having a larger model we do not suffer from severe over-fitting and perform comparably or better than top performing methods.

\subsubsection{Architecture Transferability.}
\label{subsec:mini_transfer}
It has been shown, in the case of architecture search, that it is possible to learn an architecture on a smaller dataset, e.g. CIFAR-10, and then the optimized architecture is transferable to a larger datasets, e.g. ImageNet. This helps mitigating the costly architecture search process. We follow this route, showing in our experiment that we can learn the $\hat\alpha$ values on FC100 and transfer them to miniImageNet (and then train the weights $w$ of the transferred architecture on this dataset).

For \textit{mini}ImageNet we set the $\alpha$'s to be fixed to values obtained for the FC100, while the rest of the parameters of the searched top block are randomly initialized. We find that the architecture transferred from FC100 to miniImageNet is performing well, with results comparable to other state-of-the-art methods, 62.82/79.35 for 1/5-shot. But a search performed on the actual dataset (miniImageNet training set) is outperforming the transferred one.

\subsection{Ablation studies}
\label{sec:ablation}
Next, we explore the effect of the different design choices made in our approach. 

\paragraph{Large model effect.}
We hypothesize that simply using the same algorithm with a larger model architecture would not result in better performance and it might even harm performance. This is evident in Figure \ref{fig:acc_vs_arch} when comparing the performance of different methods across increasingly larger architectures. This is also evident by observing the architectures usually used in the few-shot literature.

It is already been shown in DARTS that training $\alpha$ together with $w$ simultaneously decrease performance. They attribute this decrease to $\alpha$ over-fitting the training-set. To confirm our hypothesis, we used SGD to train our suggested DAG architecture, using fixed uniform $\alpha$ instead of the learned $\alpha$ (making it even less likely to over-fit compared to the ablation in DARTS). To this end $\alpha^{i,j}$ are initialized so each operation is given the same weight and are kept fixed. We observed that indeed in this case our large architecture is not performing as well as ResNet12 (a smaller architecture). See Table \ref{tab:ablation}b.

\begin{minipage}{0.95\textwidth}
\hspace{-0.6cm}
  \begin{minipage}{0.48\textwidth}
        \includegraphics[width=1\linewidth]{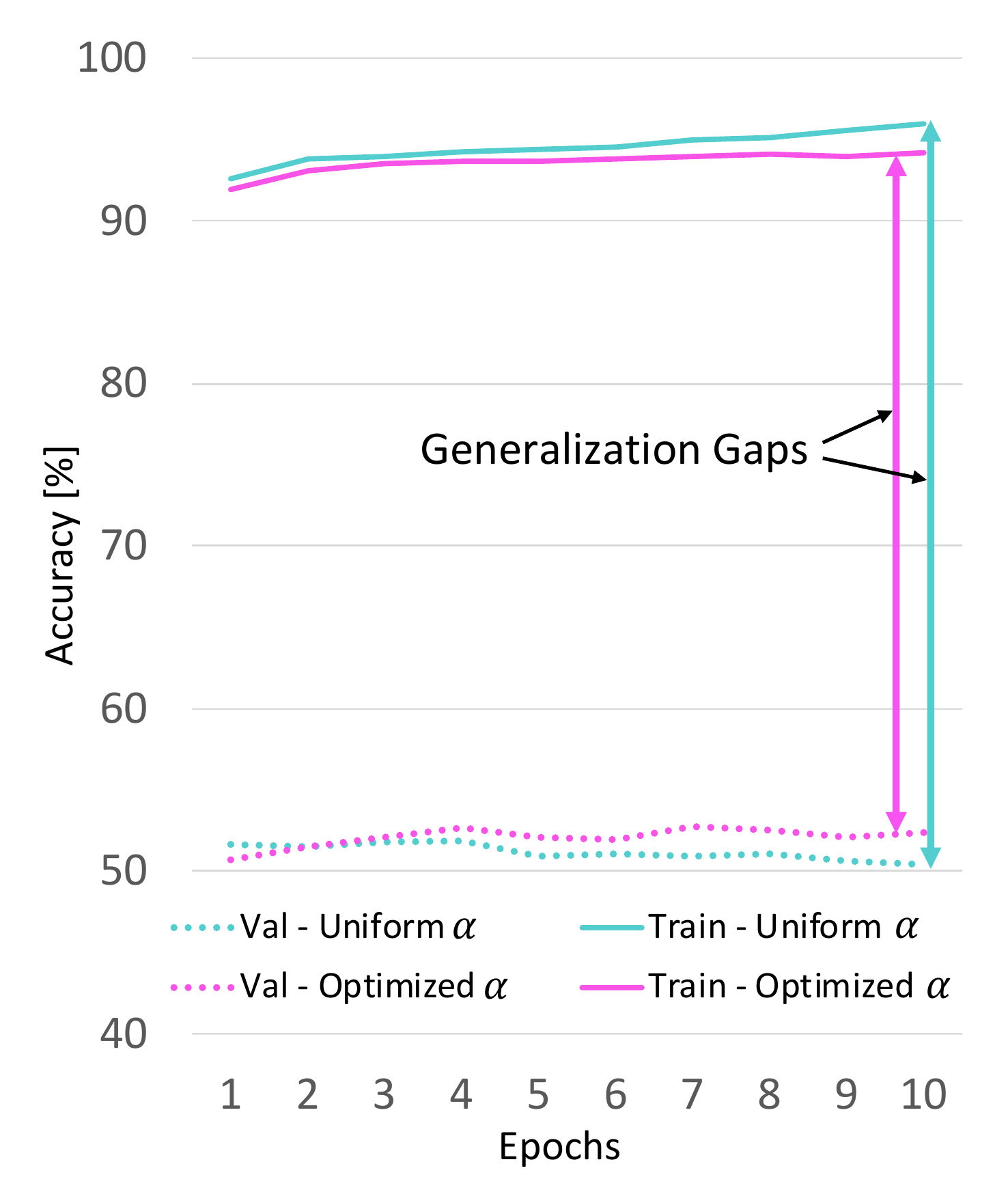}
          \captionof{figure}{\textbf{Training curves with and without optimization of} $\alpha$ for FC100. The generalization gap (gap between training-set and validation-set accuracies) is smaller when $\alpha$ is optimized using our method, suggesting it has a regularization effect. The uniform-$\alpha$ and optimized-$\alpha$ experiments are described in Sec. \ref{sec:ablation}.}
          \label{fig:generalization_gap}
  \end{minipage}
  \hfill
  \begin{minipage}{0.48\textwidth}
    \captionof{table}{\textbf{Few-shot CIFAR-100 (FC100) 5-way accuracy}}
    \centering
    \begin{tabular}{lcc}
    \toprule
     & \multicolumn{2}{c}{FC100} \\
    Method & 1-shot & 5-shot  \\
    \midrule
    ProtoNet \cite{Snell2017} & 37.50 & 52.50 \\
    TADAM \cite{tadam} & 40.10 & 56.10  \\
    MetaOptNet \cite{lee2019meta}  & 41.37 & 55.30 \\
    \midrule
    MetAdapt (Ours) & \textbf{44.83} & \textbf{58.47} \\
    \bottomrule
    \end{tabular}
    \label{tab:results-fc100}
    \vspace{1cm}
    
    \captionof{table}{\textbf{MetAdapt vs. S-MetAdapt (Stochastic MetAdapt)}; CIFAR-100 (FC100) 5-way accuracy}
    \begin{tabular}{lcc}
    \toprule
     & \multicolumn{2}{c}{FC100} \\
    Method & 1-shot & 5-shot  \\
    \midrule
    S-MetAdapt (Ours) & 41.97 & 55.31 \\
    MetAdapt (Ours) & \textbf{44.83} & \textbf{58.47} \\
    \bottomrule
    \end{tabular}
    \label{tab:smetadapt}
  \end{minipage}
\end{minipage}

\begin{table*}[tbh]
\small
\caption{\textbf{Ablation studies on FC100}}
\begin{center}
\begin{tabular}{llcccccccrr}
& Description	& \rotatebox{90}{DAG} &  \rotatebox{90}{{Optimized $\alpha$}}	&   \rotatebox{90}{{$2^{nd}$ order}}  & \rotatebox{90}{More OPs} &  \rotatebox{90}{Controllers} & \rotatebox{90}{Test augment.}& \rotatebox{90}{Test fine-tune} & 1-shot & 5-shot \\
\midrule
a. & ReseNet12    & \xmark & \xmark & \xmark	& \xmark & \xmark & \xmark & \xmark & 41.37	& 55.30 \\
b. & Replacing last ResBlock with DAG (uniform $\alpha$)	& \cmark & \xmark & \xmark	& \xmark & \xmark & \xmark & \xmark & 41.07	& 54.91 \\
c. & + Iteratively optimizing $w$ and $\alpha$	& \cmark & \cmark & \xmark	& \xmark & \xmark & \xmark & \xmark & 42.01	& 56.17 \\
d. & + $2^{nd}$ order approx. of $w$ for $\alpha$ update & \cmark & \cmark & \cmark	& \xmark & \xmark & \xmark & \xmark & 42.33	& 56.42 \\
e. & + Adding more operations ($5\times5$ ops)& \cmark & \cmark & \cmark	& \cmark & \xmark & \xmark & \xmark& 42.40	& 56.68 \\
f. & + Task-adaptivness \small{(MetAdapt Controllers)} & \cmark & \cmark & \cmark	& \cmark & \cmark & \xmark & \xmark & 44.42 & 58.54 \\
g. & + Test time flip augmentations & \cmark & \cmark & \cmark	& \cmark & \cmark & \cmark & \xmark & 44.61 & 58.49 \\
h. & + Test time fine tuning & \cmark & \cmark & \cmark	& \cmark & \cmark & \cmark & \cmark & 44.83 & 58.47 \\

\bottomrule
\end{tabular}
\end{center}
\label{tab:ablation}
\end{table*}

\paragraph{DARTS Without Meta-Adaptation.}
Next we test the effect of optimizing $\alpha$ using iterative intermittent optimization for $w$ and $\alpha$ using different folds of the training set. Here, $w$ and $\alpha$ are updated intermittently one mini-batch at a time. 
In order to see the importance of using second-order approximation of $w$ after convergence, we perform training with and without it.

\textit{Without:}
The $\alpha$ updates are done without the second-order approximation of $w$ after a gradient descent step, i.e., the updates are performed according to:
\begin{equation}
    w = w - \mu \nabla_w Loss_{train_{w}}(w,\alpha)
\end{equation}
\begin{equation}
    \alpha = \alpha  - \eta \nabla_\alpha Loss_{train_{\alpha}}(w ,\alpha).
\end{equation}
We find the $\alpha$ optimization is helping at improving the performance by about $1\%$ compared to the fixed architecture (See Table \ref{tab:ablation}c).

\textit{With: }
The $\alpha$ updates are done not according to current value of $w$ but at an approximation of its value after convergence (see Equation \ref{eq:alpha_update}). The update of $w$ is performed according to Equations \ref{eq:w_update}. We find that this change gives a moderate improvement of about $0.3\%$ (see Table \ref{tab:ablation}d).

Figure \ref{fig:generalization_gap} presents the training curves for training with the proposed bi-level optimization of $w$ and $\alpha$ vs. training the large model when $\alpha$ is fixed. It shows that the generalization gap is larger for the latter case and confirms our hypothesis that simply adding more parameters is not sufficient for good performance.

\paragraph{Number of Operations.}
In the ablation experiments described till now, we used a slightly smaller model. Each edge is composed of $6$ operations out of the $8$ listed in Table \ref{tab:operations}, excluding the $5 \times 5$ operations. Now, we add these operations to test the effect of a larger set of operations. 
Adding these operations improves slightly further the performance (see Table \ref{tab:ablation}e).

\paragraph{MetAdapt Controllers.}
Then, we add the MetAdapt Controllers, so the architecture is adapted to the current task according to the support samples. This brings us to the full MetAdapt method. We find that indeed the adaptations to each task are beneficial. The meta-adaptations improve the accuracy by around $2\%$ (see Table \ref{tab:ablation}f).

\paragraph{Test time augmentations and fine-tuning.}
Finally, we add test time flip augmentations and fine-tuning as described in \ref{sub:impl_details}.
This helps in the case of 1-shot with $+0.41\%$ improvement, but has no noticeable effect for 5-shot (see Table \ref{tab:ablation}g-h).

\paragraph{S-MetAdapt.}
A recent approach suggested for architecture search is Stochastic Neural Architecture Search (SNAS \cite{snas}). Usually for D-NAS, e.g. DARTS, at search time the training is done on the full model at each iteration where each edge is a weighted-sum of its operations according to $\alpha^{i,j}$. Contrarily, in SNAS $\alpha^{i,j}$ are treated as probabilities of a Multinomial Distribution and at each iteration a single operation is sampled accordingly. So at each iteration only a single operation per edge affects the classification outcome and only this operation is be updated in the gradient descent backward step. Of course sampling from a Multinomial Distribution directly is not differentiable, so at training time the Gumbel Distribution is used as a differentiable approximation.

We tested a SNAS version of MetAdapt, named S-MetAdapt, on the few-shot classification task. Other than the modifications specified below S-MetAdapt is similar to MetAdapt. At training time, instead of the \textit{Mixed Operation} defined in Equation \ref{eq:mixedOp}, we define the \textit{Mixed Operation} to be:
\begin{equation}\label{eq:mixedOpGumble}
\bar{o}_{i,j}(x)=\sum_{o\in\mathcal{O}}z_o^{i,j}o(x)
\end{equation}
where $z^{(i,j)}$ is a continuous approximation of a one-hot vector sampled from a Gumbel Distribution:
\begin{equation}\label{eq:Gumble}
z^{i,j} \sim Gumbel(\alpha^{i,j}).
\end{equation}
Here $\alpha^{i,j}$ are after softmax normalization and summed to $1$.
At test time, rather than the one-hot approximation, we use the operation with the top probability
\begin{equation}\label{eq:Gumble1}
z_k^{i,j} = \begin{cases}
1, & if\ k=argmax(\alpha^{i,j}) \\
0, & otherwise
\end{cases}
\end{equation}
Using this method we get better results for FC100 $1$-shot and comparable results for $5$-shot, compared to vanilla MetaOptNet. However, it does not perform as well as the non-stochastic version of MetAdapt. See Table \ref{tab:smetadapt}.

\section{Conclusions}
%
In this work we have proposed MetAdapt, a few-shot learning approach that enables meta-learned network architecture that is adaptive to novel few-shot tasks. The proposed approach effectively applies tools from the Neural Architecture Search (NAS) literature, extended with the concept of `MetAdapt Controllers', in order to learn adaptive architectures. These tools help mitigate over-fitting to the extremely small data of the few-shot tasks and domain shift between the training set and the test set. We demonstrate that the proposed approach successfully improves state-of-the-art results on two popular few-shot benchmarks, \textit{mini}ImageNet and FC100, and carefully ablate the different optimization steps and design choices of the proposed approach. 

Some interesting future work directions include extending the proposed approach to progressively searching the full network architecture (instead of just the last block), applying the approach to other few-shot tasks such as detection and segmentation, and researching into different variants of task-adaptivity including global connections modifiers and inter block adaptive wiring.

%
%
\bibliographystyle{splncs04}
\bibliography{egbib}

\begin{thebibliography}{10}
\providecommand{\url}[1]{\texttt{#1}}
\providecommand{\urlprefix}{URL }
\providecommand{\doi}[1]{https://doi.org/#1}

\bibitem{alfassy2019laso}
Alfassy, A., Karlinsky, L., Aides, A., Shtok, J., Harary, S., Feris, R.,
  Giryes, R., Bronstein, A.M.: Laso: Label-set operations networks for
  multi-label few-shot learning. In: Proceedings of the IEEE Conference on
  Computer Vision and Pattern Recognition. pp. 6548--6557 (2019)

\bibitem{Antoniou2018}
Antoniou, A., Storkey, A., Edwards, H.: {Data Augmentation Generative
  Adversarial Networks}. arXiv:1711.04340  (2017),
  \url{https://arxiv.org/pdf/1711.04340.pdf}

\bibitem{Ba2015}
Ba, J.L., Swersky, K., Fidler, S., Salakhutdinov, R.: {Predicting deep
  zero-shot convolutional neural networks using textual descriptions}.
  Proceedings of the IEEE International Conference on Computer Vision
  \textbf{2015 Inter},  4247--4255 (2015). \doi{10.1109/ICCV.2015.483}

\bibitem{proxy}
Cai, H., Zhu, L., Han, S.: Proxylessnas: Direct neural architecture search on
  target task and hardware. In: ICLR (2019)

\bibitem{closer_look}
Chen, W.Y.: {A Closer Look At Few-Shot Classification}. In: ICLR. pp. 1--16
  (2018)

\bibitem{chen2019progressive}
Chen, X., Xie, L., Wu, J., Tian, Q.: Progressive differentiable architecture
  search: Bridging the depth gap between search and evaluation. arXiv preprint
  arXiv:1904.12760  (2019)

\bibitem{Chen2018}
Chen, Z., Fu, Y., Zhang, Y., Jiang, Y.G., Xue, X., Sigal, L.: {Semantic Feature
  Augmentation in Few-shot Learning}. arXiv:1804.05298v2  (2018),
  \url{http://arxiv.org/abs/1804.05298}

\bibitem{chopra05}
Chopra, S., Hadsell, R.: {Learning a Similarity Metric Discriminatively , with
  Application to Face Verification}. CVPR  (2005)

\bibitem{Dong2017}
Dong, X., Zheng, L., Ma, F., Yang, Y., Meng, D.: {Few-shot Object Detection}.
  Arxiv:1706.08249 pp. 1--11 (2017), \url{http://arxiv.org/abs/1706.08249}

\bibitem{Dosovitskiy2017}
Dosovitskiy, A., Springenberg, J.T., Tatarchenko, M., Brox, T.: {Learning to
  Generate Chairs, Tables and Cars with Convolutional Networks}. IEEE
  Transactions on Pattern Analysis and Machine Intelligence  \textbf{39}(4),
  692--705 (2017). \doi{10.1109/TPAMI.2016.2567384}

\bibitem{Durugkar2017}
Durugkar, I., Gemp, I., Mahadevan, S.: {Generative Multi-Adversarial Networks}.
  International Conference on Learning Representations (ICLR) pp. 1--14 (2017).
  \doi{10.1016/j.ajodo.2005.02.022}, \url{http://arxiv.org/abs/1611.01673}

\bibitem{Dvornik_2019_ICCV}
Dvornik, N., Schmid, C., Mairal, J.: Diversity with cooperation: Ensemble
  methods for few-shot classification. In: The IEEE International Conference on
  Computer Vision (ICCV) (October 2019)

\bibitem{Finn2017}
Finn, C., Abbeel, P., Levine, S.: {Model-Agnostic Meta-Learning for Fast
  Adaptation of Deep Networks}. arXiv:1703.03400  (2017),
  \url{http://arxiv.org/abs/1703.03400}

\bibitem{Fu2015}
Fu, Y., Hospedales, T.M., Xiang, T., Gong, S.: {Transductive Multi-View
  Zero-Shot Learning}. IEEE Transactions on Pattern Analysis and Machine
  Intelligence  \textbf{37}(11),  2332--2345 (2015).
  \doi{10.1109/TPAMI.2015.2408354}

\bibitem{Fu2016a}
Fu, Y., Sigal, L.: {Semi-supervised Vocabulary-informed Learning}. IEEE
  Conference on Computer Vision and Pattern Recognition (CVPR) pp. 5337--5346
  (2016). \doi{10.1109/CVPR.2016.576}

\bibitem{Garcia2017}
Garcia, V., Bruna, J.: {Few-Shot Learning with Graph Neural Networks}.
  arXiv:1711.04043 pp. 1--13 (2017), \url{http://arxiv.org/abs/1711.04043}

\bibitem{gidaris2019boosting}
Gidaris, S., Bursuc, A., Komodakis, N., P{\'e}rez, P., Cord, M.: Boosting
  few-shot visual learning with self-supervision. In: Proceedings of the IEEE
  International Conference on Computer Vision (2019)

\bibitem{gidaris2018dynamic}
Gidaris, S., Komodakis, N.: Dynamic few-shot visual learning without
  forgetting. In: Proceedings of the IEEE Conference on Computer Vision and
  Pattern Recognition. pp. 4367--4375 (2018)

\bibitem{gidaris2019generating}
Gidaris, S., Komodakis, N.: Generating classification weights with gnn
  denoising autoencoders for few-shot learning. In: Proceedings of the IEEE
  Conference on Computer Vision and Pattern Recognition. pp. 21--30 (2019)

\bibitem{Goodfellow2014}
Goodfellow, I., Pouget-Abadie, J., Mirza, M., Xu, B., Warde-Farley, D., Ozair,
  S., Courville, A., Bengio, Y.: {Generative Adversarial Nets}. Advances in
  Neural Information Processing Systems 27 pp. 2672--2680 (2014).
  \doi{10.1017/CBO9781139058452},
  \url{http://papers.nips.cc/paper/5423-generative-adversarial-nets.pdf}

\bibitem{Guu2017}
Guu, K., Hashimoto, T.B., Oren, Y., Liang, P.: {Generating Sentences by Editing
  Prototypes}. Arxiv:1709.08878  (2017),
  \url{https://arxiv.org/pdf/1709.08878.pdf}

\bibitem{Hariharan2017}
Hariharan, B., Girshick, R.: {Low-shot Visual Recognition by Shrinking and
  Hallucinating Features}. IEEE International Conference on Computer Vision
  (ICCV)  (2017), \url{https://arxiv.org/pdf/1606.02819.pdf}

\bibitem{Huang2017}
Huang, G., Liu, Z., v.~d. Maaten, L., Weinberger, K.Q.: {Densely Connected
  Convolutional Networks}. 2017 IEEE Conference on Computer Vision and Pattern
  Recognition (CVPR) pp. 2261--2269 (2017). \doi{10.1109/CVPR.2017.243},
  \url{https://arxiv.org/pdf/1608.06993.pdf}

\bibitem{caml}
Jiang, X., Havaei, M., Varno, F., Chartrand, G., Chapados, N., Matwin, S.:
  Learning to learn with conditional class dependencies. ICLR  (2018)

\bibitem{Zhu2016}
Jun-Yan~Zhu, Philipp~Krahenbuhl, E.S., Efros, A.: {Generative Visual
  Manipulation on the Natural Image Manifold}. European Conference on Computer
  Vision (ECCV). pp. 597--613 (2016)

\bibitem{kim2018auto}
Kim, J., Lee, S., Kim, S., Cha, M., Lee, J.K., Choi, Y., Choi, Y., Cho, D.Y.,
  Kim, J.: Auto-meta: Automated gradient based meta learner search. arXiv
  preprint arXiv:1806.06927  (2018)

\bibitem{Krizhevsky2009}
Krizhevsky, A.: {Learning Multiple Layers of Features from Tiny Images}.
  Technical report. Science Department, University of Toronto, Tech. pp. 1--60
  (2009). \doi{10.1.1.222.9220}

\bibitem{Krizhevsky2012}
Krizhevsky, A., Sutskever, I., Hinton, G.E.: {ImageNet Classification with Deep
  Convolutional Neural Networks}. Advances In Neural Information Processing
  Systems pp.~1--9 (2012). \doi{http://dx.doi.org/10.1016/j.protcy.2014.09.007}

\bibitem{lee2019meta}
Lee, K., Maji, S., Ravichandran, A., Soatto, S.: Meta-learning with
  differentiable convex optimization. In: CVPR (2019)

\bibitem{Li2017}
Li, Z., Zhou, F., Chen, F., Li, H.: {Meta-SGD: Learning to Learn Quickly for
  Few-Shot Learning}. arXiv:1707.09835  (2017),
  \url{http://arxiv.org/abs/1707.09835}

\bibitem{Li2016}
Li, Z., Hoiem, D.: {Learning without Forgetting}. IEEE Transactions on Pattern
  Analysis and Machine Intelligence pp. 1--13 (2016).
  \doi{10.1007/978-3-319-46493-0{\_}37}, \url{http://arxiv.org/abs/1606.09282}

\bibitem{Lim2012}
Lim, J.J., Salakhutdinov, R., Torralba, A.: {Transfer Learning by Borrowing
  Examples for Multiclass Object Detection}. Advances in Neural Information
  Processing Systems 26 (NIPS) pp.~1--9 (2012)

\bibitem{liu2018progressive}
Liu, C., Zoph, B., Neumann, M., Shlens, J., Hua, W., Li, L.J., Fei-Fei, L.,
  Yuille, A., Huang, J., Murphy, K.: Progressive neural architecture search.
  In: Proceedings of the European Conference on Computer Vision (ECCV). pp.
  19--34 (2018)

\bibitem{liu19darts}
Liu, H., Simonyan, K., Yang, Y.: Darts: Differentiable architecture search. In:
  International Conference on Learning Representations (ICLR) (2019)

\bibitem{Mao2016}
Mao, X., Li, Q., Xie, H., Lau, R.Y.K., Wang, Z., Smolley, S.P.: {Least Squares
  Generative Adversarial Networks}. IEEE International Conference on Computer
  Vision (ICCV) pp. 1--16 (2016). \doi{10.1109/ICCV.2017.304},
  \url{http://arxiv.org/abs/1611.04076}

\bibitem{Mishra2017AMeta-Learner}
Mishra, N., Rohaninejad, M., Chen, X., Abbeel, P.: {A Simple Neural Attentive
  Meta-Learner}. Advances In Neural Information Processing Systems (NIPS)
  (2017), \url{https://arxiv.org/pdf/1707.03141.pdf}

\bibitem{Munkhdalai2017}
Munkhdalai, T., Yu, H.: {Meta Networks}. arXiv:1703.00837  (2017).
  \doi{10.1093/mnrasl/slx008}, \url{http://arxiv.org/abs/1703.00837}

\bibitem{munkhdalai2018rapid}
Munkhdalai, T., Yuan, X., Mehri, S., Trischler, A.: Rapid adaptation with
  conditionally shifted neurons. In: International Conference on Machine
  Learning. pp. 3661--3670 (2018)

\bibitem{nichol2018first}
Nichol, A., Achiam, J., Schulman, J.: On first-order meta-learning algorithms.
  arXiv preprint arXiv:1803.02999  (2018)

\bibitem{ASAP}
Noy, A., Nayman, N., Ridnik, T., Zamir, N., Doveh, S., Friedman, I., Giryes,
  R., Zelnik-Manor, L.: Asap: Architecture search, anneal and prune. In:
  arXiv:1904.04123 (2019)

\bibitem{tadam}
Oreshkin, B.N., Rodriguez, P., Lacoste, A.: {TADAM: Task dependent adaptive
  metric for improved few-shot learning}. NeurIPS  (5 2018),
  \url{http://arxiv.org/abs/1805.10123}

\bibitem{Park2015}
Park, D., Ramanan, D.: {Articulated pose estimation with tiny synthetic
  videos}. IEEE Conference on Computer Vision and Pattern Recognition (CVPR)
  \textbf{2015-Octob},  58--66 (2015). \doi{10.1109/CVPRW.2015.7301337}

\bibitem{ENAS}
Pham, H., Y~Guan, M., Zoph, B., V.~Le, Q., , Dean, J.: Efficient neural
  architecture search via parameter sharing. In: International Conference on
  Machine Learning (ICML) (2018)

\bibitem{qiao2018few}
Qiao, S., Liu, C., Shen, W., Yuille, A.L.: Few-shot image recognition by
  predicting parameters from activations. In: Proceedings of the IEEE
  Conference on Computer Vision and Pattern Recognition. pp. 7229--7238 (2018)

\bibitem{Radford2015}
Radford, A., Metz, L., Chintala, S.: {Unsupervised Representation Learning with
  Deep Convolutional Generative Adversarial Networks}. arXiv:1511.06434 pp.
  1--16 (2015). \doi{10.1051/0004-6361/201527329},
  \url{http://arxiv.org/abs/1511.06434}

\bibitem{Ravi2017}
Ravi, S., Larochelle, H.: {Optimization As a Model for Few-Shot Learning}. ICLR
  pp. 1--11 (2017), \url{https://openreview.net/pdf?id=rJY0-Kcll}

\bibitem{Real18Regularized}
Real, E., Aggarwal, A., Huang, Y., V.~Le, Q.: Regularized evolution for image
  classifier architecture search. In: International Conference on Machine
  Learning - ICML AutoML Workshop (2018)

\bibitem{Reed2018}
Reed, S., Chen, Y., Paine, T., van~den Oord, A., Eslami, S.M.A., Rezende, D.,
  Vinyals, O., de~Freitas, N.: {Few-shot autoregressive density estimation:
  towards learning to learn distributions}. arXiv:1710.10304  (2018)

\bibitem{Rippel2015}
Rippel, O., Paluri, M., Dollar, P., Bourdev, L.: {Metric Learning with Adaptive
  Density Discrimination}. arXiv:1511.05939 pp. 1--15 (2015),
  \url{http://arxiv.org/abs/1511.05939}

\bibitem{imagenet}
Russakovsky, O., Deng, J., Su, H., Krause, J., Satheesh, S., Ma, S., Huang, Z.,
  Karpathy, A., Khosla, A., Bernstein, M., Berg, A.C., Fei-Fei, L.: {ImageNet
  Large Scale Visual Recognition Challenge}. IJCV  (9 2015),
  \url{http://arxiv.org/abs/1409.0575}

\bibitem{leo}
Rusu, A.A., Rao, D., Sygnowski, J., Vinyals, O., Pascanu, R., Osindero, S.,
  Hadsell, R.: {Meta-Learning with Latent Embedding Optimization}. In: ICLR (7
  2018), \url{http://arxiv.org/abs/1807.05960}

\bibitem{Santoro2016}
Santoro, A., Bartunov, S., Botvinick, M., Wierstra, D., Lillicrap, T.:
  {Meta-Learning with Memory-Augmented Neural Networks}. Journal of Machine
  Learning Research  \textbf{48}(Proceedings of The 33rd International
  Conference on Machine Learning),  1842--1850 (2016).
  \doi{10.1002/2014GB005021}

\bibitem{schwartz2019baby}
Schwartz, E., Karlinsky, L., Feris, R., Giryes, R., Bronstein, A.M.: Baby steps
  towards few-shot learning with multiple semantics. arXiv preprint
  arXiv:1906.01905  (2019)

\bibitem{Schwartz2018}
Schwartz, E., Karlinsky, L., Shtok, J., Harary, S., Marder, M., Kumar, A.,
  Feris, R., Giryes, R., Bronstein, A.M.: {Delta-Encoder: an Effective Sample
  Synthesis Method for Few-Shot Object Recognition}. NIPS  (2018),
  \url{https://arxiv.org/pdf/1806.04734.pdf}

\bibitem{Snell2017}
Snell, J., Swersky, K., Zemel, R.: {Prototypical Networks for Few-shot
  Learning}. NIPS  (2017), \url{http://arxiv.org/abs/1703.05175}

\bibitem{Su2015}
Su, H., Qi, C.R., Li, Y., Guibas, L.J.: {Render for CNN Viewpoint Estimation in
  Images Using CNNs Trained with Rendered 3D Model Views.pdf}. IEEE
  International Conference on Computer Vision (ICCV) pp. 2686--2694 (2015)

\bibitem{Sung2017LearningLearning}
Sung, F., Yang, Y., Zhang, L., Xiang, T., Torr, P.H.S., Hospedales, T.M.:
  {Learning to Compare: Relation Network for Few-Shot Learning}.
  arXiv:1711.06025  (11 2017), \url{http://arxiv.org/abs/1711.06025}

\bibitem{relationnet}
Sung, F., Yang, Y., Zhang, L., Xiang, T., Torr, P.H., Hospedales, T.M.:
  Learning to compare: Relation network for few-shot learning. In: Proceedings
  of the IEEE Conference on Computer Vision and Pattern Recognition. pp.
  1199--1208 (2018)

\bibitem{effnet}
Tan, M., Le, Q.V.: Efficientnet: Rethinking model scaling for convolutional
  neural networks. arXiv preprint arXiv:1905.11946  (2019)

\bibitem{Vinyals2016}
Vinyals, O., Blundell, C., Lillicrap, T., Kavukcuoglu, K., Wierstra, D.:
  {Matching Networks for One Shot Learning}. NIPS  (2016).
  \doi{10.1109/CVPR.2016.95}, \url{http://arxiv.org/abs/1606.04080}

\bibitem{Wang2018}
Wang, Y.X., Girshick, R., Hebert, M., Hariharan, B.: {Low-Shot Learning from
  Imaginary Data}. arXiv:1801.05401  (2018),
  \url{http://arxiv.org/abs/1801.05401}

\bibitem{Wang2016a}
Wang, Y.X., Hebert, M.: {Learning from Small Sample Sets by Combining
  Unsupervised Meta-Training with CNNs}. Advances In Neural Information
  Processing Systems (NIPS) pp. 244--252 (2016),
  \url{https://papers.nips.cc/paper/6408-learning-from-small-sample-sets-by-combining-unsupervised-meta-training-with-cnns}

\bibitem{Wang2016}
Wang, Y.X., Hebert, M.: {Learning to Learn: Model Regression Networks for Easy
  Small Sample Learning}. European Conference on Computer Vision (ECCV) pp.
  616--634 (2016). \doi{10.1007/978-3-319-46466-4}

\bibitem{Weinberger2009}
Weinberger, K.Q., Saul, L.K.: {Distance Metric Learning for Large Margin
  Nearest Neighbor Classification}. The Journal of Machine Learning Research
  \textbf{10},  207--244 (2009). \doi{10.1126/science.277.5323.215}

\bibitem{snas}
Xie, S., Zheng, H., Liu, C., Lin, L.: Snas: Stochastic neural architecture
  search. In: International Conference on Learning Representations (ICLR) (12
  2018)

\bibitem{am3}
Xing, C., Rostamzadeh, N., Oreshkin, B.N., Pinheiro, P.O.: {Adaptive
  Cross-Modal Few-Shot Learning}. Arxiv  (2019),
  \url{https://arxiv.org/pdf/1902.07104.pdf}

\bibitem{Yu2017}
Yu, A., Grauman, K.: {Semantic Jitter: Dense Supervision for Visual Comparisons
  via Synthetic Images}. Proceedings of the IEEE International Conference on
  Computer Vision  \textbf{2017-Octob},  5571--5580 (2017).
  \doi{10.1109/ICCV.2017.594}

\bibitem{Zhou2018}
Zhou, F., Wu, B., Li, Z.: {Deep Meta-Learning: Learning to Learn in the Concept
  Space}. arXiv:1802.03596  (2 2018), \url{http://arxiv.org/abs/1802.03596}

\bibitem{Zhu2017}
Zhu, J.Y., Park, T., Isola, P., Efros, A.A.: {Unpaired Image-to-Image
  Translation Using Cycle-Consistent Adversarial Networks}. Proceedings of the
  IEEE International Conference on Computer Vision  \textbf{2017-Octob},
  2242--2251 (2017). \doi{10.1109/ICCV.2017.244}

\bibitem{zophNasRL}
Zoph, B., V.~Le, Q.: Neural architecture search with reinforcement learning.
  In: International Conference on Learning Representations (ICLR) (2017)

\end{thebibliography}
\end{document}